%% file: nips2025_position_paper.tex
\documentclass{article}

\PassOptionsToPackage{numbers, compress}{natbib}

\usepackage[preprint]{neurips_2025}

\usepackage{xurl}           
\usepackage[utf8]{inputenc} 
\usepackage[T1]{fontenc}    
\usepackage{hyperref}       
\usepackage{url}            
\usepackage{booktabs}       
\usepackage{amsfonts}       
\usepackage{nicefrac}       
\usepackage{microtype}      
\usepackage{xcolor}         
\usepackage{graphicx}       
\usepackage{subcaption}
\usepackage{amsmath}
\usepackage{amssymb}
\usepackage{mathtools}
\usepackage{amsthm}
\usepackage{enumitem} 

\usepackage[capitalize,noabbrev]{cleveref}

\usepackage{tcolorbox}      
\usepackage{tikz}           
\usetikzlibrary{arrows.meta, positioning}
\usepackage{algorithm}      
\usepackage{algpseudocode}  

\theoremstyle{plain}

\theoremstyle{definition}

\theoremstyle{remark}

\title{The Missing Reward: Active Inference in the Era of Experience}

\author{
Bo Wen \\
IBM T.J. Watson Research Center \\
Yorktown Heights, NY \\
\texttt{bwen@us.ibm.com} \\
}

\begin{document}

\maketitle

\begin{abstract}
This paper argues that Active Inference (AIF) provides a crucial foundation for developing autonomous AI agents capable of learning from experience without continuous human reward engineering. As AI systems begin to exhaust high-quality training data and rely on increasingly large human workforces for reward design, the current paradigm faces significant scalability challenges that could impede progress toward genuinely autonomous intelligence. The proposal for an ``Era of Experience,'' where agents learn from self-generated data, is a promising step forward. However, this vision still depends on extensive human engineering of reward functions, effectively shifting the bottleneck from data curation to reward curation. This highlights what we identify as the \textbf{grounded-agency gap}: the inability of contemporary AI systems to autonomously formulate, adapt, and pursue objectives in response to changing circumstances. We propose that AIF can bridge this gap by replacing external reward signals with an intrinsic drive to minimize free energy, allowing agents to naturally balance exploration and exploitation through a unified Bayesian objective. By integrating Large Language Models as generative world models with AIF's principled decision-making framework, we can create agents that learn efficiently from experience while remaining aligned with human values. This synthesis offers a compelling path toward AI systems that can develop autonomously while adhering to both computational and physical constraints.
\end{abstract}

\section{Introduction}
\label{sec:introduction}

In his NeurIPS 2024 award speech, Ilya Sutskever declared that ``pre-training as we know it will unquestionably end'' \cite{sutskever2024pretraining}. Dario Amodei, in a recent podcast\cite{amodei2023concerns}, estimated a ``10\% chance that the scaling of AI systems could stagnate due to insufficient data''. Several empirical studies \cite{villalobos2024will, chinchillaswild2022} support similar conclusions and project that demand for training data will soon exceed the available public supply. As high-quality human-generated data becomes increasingly scarce, new approaches must emerge to sustain AI's advancement.

Addressing this looming data shortage, the recent preprint ``Welcome to the Era of Experience'' (abbreviated as ``EoE'' in the following) \cite{silver2025era} by Silver and Sutton proposes a paradigm shift from static, human-generated datasets to dynamic, agent-generated experiences. We interpret this as AI agents learning continuously through their own environmental interactions, effectively creating a self-sustaining cycle where these interactions generate the very data needed for ongoing training and improvement. This closed-loop system not only circumvents concerns about exhausting training data supplies but potentially unlocks capabilities beyond what human-curated data alone could achieve.

EoE extends far beyond conventional reinforcement learning, outlining several essential components for this experiential paradigm. Silver and Sutton envision agents that inhabit continuous streams of experience rather than isolated interaction episodes, with actions and observations richly grounded in their environments. These agents would \textit{derive rewards from real-world signals instead of human judgments}, and employ planning mechanisms that reason directly about experience rather than abstract concepts.

However, their blueprint leaves open a practical, and potentially decisive, question. The challenge of determining who will engineer the reward functions that turn raw signals into useful guidance remains unresolved. In their own examples, different reward functions must be tuned for each high-level user goal. At the scale implied by lifelong, open-ended agents, this merely shifts the bottleneck from curating training data to curating reward functions, reintroducing the kind of domain knowledge engineering that Sutton's \emph{bitter lesson} warns us about \cite{sutton2019bitter}.

\textbf{We argue that Active Inference (AIF) provides the missing foundation for autonomous AI agents that can learn from experience without constant human reward engineering.} By replacing external reward engineering with intrinsic free energy minimization, AIF agents naturally balance exploration and exploitation through a unified Bayesian objective that emerges from the agent's own world model and preferences. This approach offers a more direct path to the experiential learning paradigm envisioned by Silver and Sutton, with advantages in scalability, safety, and efficiency that we explore throughout this paper.

\paragraph{Overview of Core Arguments}
This paper identifies a fundamental ``grounded-agency gap'' in contemporary AI—the inability to autonomously form, evaluate, and adapt objectives—that persists even in proposed experience-driven paradigms. We argue that Active Inference provides a compelling theoretical foundation for the ``Era of Experience'' by offering intrinsic motivation through free energy minimization, which can eliminate the need for continuous reward engineering. Building on this, we propose a novel integration where Large Language Models serve as learned generative world models within an Active Inference decision-making framework, combining the scalability of modern deep learning with the theoretical rigor of the Free Energy Principle. Finally, we situate this proposal within the physical constraints of AI development, arguing that the energy efficiency of free energy minimization is not just computationally advantageous but may be a thermodynamic necessity for sustainable AI progress.

\section{Two Dark Clouds over Contemporary AI}\label{sec:limits}

In 1900, Lord Kelvin observed that physics seemed nearly complete save for "two small clouds" on the horizon. These clouds (blackbody radiation and the Michelson-Morley experiment) ultimately revolutionized our understanding through quantum mechanics and relativity. Today's AI faces analogous clouds that signal not minor adjustments but fundamental limitations in our current paradigm.

\paragraph{Cloud I: Resource Saturation—Physical Limits of Scale.} The prevailing wisdom in AI has been simple: more data and more compute yield better models. This scaling hypothesis drove remarkable progress from early language models to today's trillion-parameter systems. However, we are reaching hard physical limits on multiple fronts simultaneously. On the data side, high-quality human text accumulated over centuries is being rapidly depleted, with quality English text projected for exhaustion within a decade while specialized domains already face severe scarcity. Web crawls yield diminishing returns: more spam, more synthetic content, less genuine human knowledge. On the compute side, training costs rise super-linearly with model size while universities and smaller labs are priced out of frontier research~\cite{cmswire2024universities}, creating unprecedented concentration of power. This is not merely an economic problem; it is a thermodynamic one. Each doubling of model capability demands exponentially more energy, while performance gains follow a logarithmic curve. This creates a fundamental mismatch that no amount of engineering can overcome.

\paragraph{Cloud II: Externalized Cognition—Hidden Human Dependencies.} The second constraint is more subtle but equally limiting: today's ``autonomous'' AI systems depend on vast networks of human cognition operating behind the scenes. This represents a fundamental architectural limitation where judgment, adaptation, and error correction are outsourced to human workers. Content moderators suffer psychological trauma from filtering toxic outputs, while annotator agreement rates reveal instability in the foundation of ``ground truth.'' More critically, this dependency scales \emph{inversely} with capability: as models grow more sophisticated, they require more nuanced human judgment for alignment, creating an ever-expanding need for specialized expertise. The recent DeepMind study on preference drift \cite{deepmind2024nonstationary} exposes the fundamental challenge: human values are not static targets but dynamic, context-dependent processes that no amount of labeling can fully capture. The result is a Sisyphean cycle: engineers continuously patch reward functions, annotators endlessly refine preference datasets, and safety teams perpetually chase emerging failure modes. This is not intelligence; it is an elaborate puppet show where humans operate the strings from behind an algorithmic curtain, just as Fei-Fei Li said, ``There's nothing artificial about artificial intelligence.'' \cite{dailyprincetonian2024fei}

\begin{table}[h]
\centering
\caption{Evidence of Structural Bottlenecks in Contemporary AI}
\label{tab:bottlenecks}
\begin{tabular}{p{0.12\textwidth} p{0.26\textwidth} p{0.26\textwidth} p{0.24\textwidth}}
\toprule
\textbf{Bottleneck} & \textbf{Resource Saturation} & \textbf{Externalized Cognition} & \textbf{Systemic Impact} \\
\midrule
\textbf{Data Scarcity} & High-quality text exhausted within 10 years~\cite{villalobos2024will}. Diminishing returns at trillion-token scales~\cite{chinchillaswild2022}. & Human-labeled data <70\% inter-annotator agreement~\cite{bai2022traininghelpfulharmlessassistant}. Quality ceiling from noisy judgments. & Progress tied to finite historical accumulation; innovation stalls without new sources \\
\midrule
\textbf{Compute / Energy} & SOTA training: 1000s GPUs, 4-5 digit tonne CO$_2$~\cite{gpt3_emissions}. Google: 48\% data center emission jump~\cite{google_datacenter, google2023sustainability}. & RLHF workforce scales with model capability~\cite{grey2023hidden}. Human oversight costs compound with AI sophistication. & Only tech giants compete~\cite{mitsloan2023industry}; universities priced out~\cite{stanfordhai2024expanding}. Environmental costs threaten social license.\\
\midrule
\textbf{Human \newline Labor} & OpenAI spends \$100M+ annually on RLHF contractors. Workers earn below living wages~\cite{hara2018earnings}. & Invisible supply chain of labelers, moderators. Psychological trauma from disturbing content~\cite{grey2023hidden}. & ``Automation'' paradox: AI creates more human work. Undermines autonomous intelligence narrative. \\
\midrule
\textbf{Economic Friction} & <10\% AI startups profitable (2024). Job displacement in automatable sectors~\cite{acemoglu2020robots}. & Cultural/linguistic bias: AI works best for dominant languages. Global inequities reinforced. & Social backlash from displacement. Talent drain from ethics concerns. Regulatory momentum building. \\
\bottomrule
\end{tabular}
\end{table}

These four symptoms (data exhaustion, compute barriers, labor dependencies, and economic friction) are not independent failures but manifestations of the two fundamental clouds. Resource saturation (Cloud I) drives the data and compute crises, while externalized cognition (Cloud II) creates the human labor and economic imbalances. Together, they reveal that scaling alone cannot produce genuine intelligence.

\section{The Grounded-Agency Gap} \label{sec:agency_gap}
These system-level constraints reveal a deeper problem: most contemporary AI systems lack the ability to autonomously create, update, and pursue objectives as circumstances change - a deficiency we term the \textbf{grounded-agency gap}. To close this gap, an agent must (i) perceive its situation, (ii) revise its goals in response to evolving user needs and environments, and (iii) operate safely without requiring continuous human intervention. In the following discussion, we demonstrate why two prevalent approaches - reward engineering and self-play - fail to achieve true agency.

\paragraph{Reward Engineering is \emph{not} Grounded Agency.} \label{sec:reward_vs_agency}
The ``reward-is-enough'' idea \cite{silver2021reward} suggests that a single, well-designed reward signal could produce all intelligent behaviors if maximized in a sufficiently complex environment. Recent successes in robotic research supported this insight: RT-2 handles kitchen tools from vision alone \cite{brohan2023rt2}, RoboCat shows impressive adaptability by fine-tuning itself to dozens of new tasks \cite{rogers2023robocat}, and PaLM-E combines language with physical actions \cite{driess2023palme}. While impressive, these systems operate in highly controlled experimental environments with clear, measurable rewards perfectly matched to specific tasks. This controlled setting masks several critical limitations when considering true grounded agency in open-world scenarios: (1) \textbf{Artificial feedback systems.} Concepts like ``tasty food'' or ``safe move'' are not directly observable. Current systems need special sensors, training resets, or expensive simulators like Habitat-2.0 \cite{savva2023habitat} and iGibson \cite{fang2020igibson} to provide constant reward feedback. This infrastructure does not exist in the real world. (2) \textbf{Short-term, fixed goals.} These demonstrations show robots performing brief tasks like pouring drinks or folding clothes. They do not handle long-term changes like shifting food preferences or unexpected safety issues that arise in real homes. The systems cannot adapt over time or learn from past experience when goals change. (3) \textbf{Hidden guard-rails and implicit constraints.} To prevent failures, engineers add safety features that are not part of the official reward. These include collision avoidance, penalty systems, or human monitors. These features shape the robot's behavior but are not part of its explicit objectives.

Consider a toy example that illustrates why this approach fails. An autonomous lab assistant is tasked with conducting a fluorescent enzyme assay and instructed to ``complete the assay efficiently.'' When the assistant observes unexpected acidification (yellow indicator), its primary reward signal pushes toward rapid completion. Without explicit safety constraints programmed into the reward function, the agent might quickly add concentrated base to neutralize the acid, potentially causing dangerous splashing of corrosive reagents or thermal damage to heat-sensitive enzymes. The original efficiency command remains unchanged even as new safety hazards emerge.

Engineers might respond by adding penalty terms for ``chemical spills'' or ``temperature violations,'' but this requires anticipating every possible failure mode and manually encoding safety preferences—a combinatorially explosive engineering challenge. Supporters might argue the system would learn safer protocols after experiencing negative outcomes, but this approach necessitates either dangerous real-world experimentation or expensive high-fidelity simulations that may not capture all safety-critical dynamics. Each new domain or task variation demands fresh rounds of reward engineering, creating the very labor-intensive bottleneck that autonomous systems should eliminate.

Under Active Inference, safety preferences like ``avoid spilling corrosive reagents'' and ``maintain enzyme integrity'' sit directly in the \textit{C} matrix as intrinsic preferences rather than external constraints. The agent naturally selects policies that minimize expected free energy by both resolving uncertainty (measuring pH first) and satisfying safety preferences (careful titration) without requiring explicit reward engineering for each potential hazard.

Recent studies from DeepMind show that this problem is real rather than merely theoretical: even with sophisticated preference learning techniques like Non-Stationary Direct Preference Optimization, user preferences exhibit significant temporal drift, causing standard algorithms to become misaligned over time \cite{deepmind2024nonstationary}. The labor-intensive nature of collecting and curating human preferences for RLHF, with its inherent inconsistencies and costs (as discussed in Section~\ref{sec:limits}), further highlights the unsustainability of continuous external reward adjustment. \textbf{Take-away:} Current reward engineering practice incurs substantial \emph{cost} (instrumentation, supervision, post-hoc patching) but rarely instills robust, general-purpose \emph{capability}. These approaches fix problems after they happen instead of giving AI the ability to adapt safely to new situations. This creates a paradox: we want autonomous AI, but we keep needing humans to constantly adjust its rewards, just with different job titles like ``reward engineer'' or ``ground-signal curator.'' The costs keep growing as AI gets more powerful.

\paragraph{Why Self-Play Produces Only Simulated Agency.} \label{sec:simulated_agency}
The challenges of open-world reward engineering stand in stark contrast to the domain-specific successes of systems like AlphaZero \cite{silver2018general} and AlphaProof \cite{romera2024mathematical}. These systems achieve superhuman performance in complex games (chess, shogi, and Go) and mathematical theorem proving through self-play, requiring minimal human input beyond the fundamental rules and clear success criteria (win conditions or valid proofs). They demonstrate remarkable ability to develop strategies and discover knowledge that exceeds human expertise, apparently without continuous external guidance or reward shaping.

However, these accomplishments represent what we call \emph{simulated agency}. They function within precisely defined, closed environments where judgment is externalized into fixed rules, explicit win conditions, and stable objectives. The environment serves as a perfect, unambiguous success oracle. When moving from these structured domains to the complexities of open-world robotics or general-purpose AI assistants, these foundational assumptions break down: (1) \textbf{Fuzzy, multifaceted objectives:} Real-world tasks rarely have a single, easily quantifiable win condition. Goals are often ill-defined, composed of multiple potentially conflicting sub-goals, and subject to interpretation (e.g., ``be a helpful assistant,'' ``ensure user well-being''). (2) \textbf{Preferences drift and evolve:} Human preferences change over time due to learning, shifting circumstances, or developing tastes. Agents relying on fixed, externally defined rewards cannot accommodate these changes. (3) \textbf{Non-stationary environments:} The real world is constantly changing in unpredictable ways. Unlike game boards with fixed rules, AI agents must handle novel situations, new entities, and evolving causal relationships. (4) \textbf{Ambiguous, sparse feedback:} Clear reward signals are rare. Environmental or user feedback is often delayed, noisy, incomplete, sometimes even contradictory. All requires substantial interpretation.

In such open-ended settings, the reward function must be \emph{learned, inferred, or adapted} rather than simply provided. While self-play excels when evaluative criteria are fixed and externally supplied, it lacks any intrinsic mechanism for agents to \emph{derive or update} these criteria from its experiences in an ambiguous, evolving world. Therefore, for general AI to achieve the self-improvement and emergent capabilities demonstrated by AlphaZero, it requires more than self-play. It needs genuine agency: the capacity to form, evaluate, and refine objectives and understanding based on its interactions with complex, underspecified environments. The fundamental challenge involves enabling AI to interpret human guidance and environmental signals as flexible preferences and evidence rather than rigid rewards, then autonomously adapt its capabilities to satisfy these evolving preferences in novel situations.

\paragraph{Bridge to Active Inference.} \label{sec:bridge}
Silver and Sutton's proposal of an ``adaptive reward, based on grounded signals, guided by user'' shares a conceptual similarity with our argument for grounded agency. Both approaches recognize the need for AI agents to dynamically adjust their objectives based on real-world interactions rather than relying on static, pre-defined rewards. However, the critical distinction lies in the \textit{locus of control}: while Sutton's framework still requires external human guidance to adapt reward functions, our approach seeks to internalize this process entirely.

This mirrors human development: just as children ``grow up'' by generalizing from lessons to broader principles, we need AI systems to develop intrinsic mechanisms for self-directed learning. The key insight is that we need a \textit{universal learning signal} that enables agents to abstract meta-level knowledge from experience, with autonomy as the implicit goal. This signal must be: intrinsically computable (derivable from the agent's sensory stream), domain-agnostic (applicable across diverse environments), preference-sensitive (responsive to human values without rigid specification), and exploration-aware (naturally balancing knowledge-seeking and goal-directed behavior).

This is precisely where the \textbf{Free Energy Principle} from Active Inference provides a principled answer. Rather than engineering countless reward functions, AIF offers a single, universal objective (minimizing expected free energy) that naturally gives rise to intelligent behavior. The agent's exploration, learning, and goal-pursuit emerge from this unified principle, making the ``Era of Experience'' both theoretically grounded and practically achievable.

\section{Active Inference with Language Models: A Path to Grounded Agency}\label{sec:active_inference_framework}

Section~\ref{sec:limits} highlighted a critical grounded-agency gap: contemporary AI, even in the envisioned ``Era of Experience,'' lacks an intrinsic mechanism for self-directed learning and judgment, often falling back on external reward engineering. To address this fundamental challenge, we need a framework that can provide AI systems with principled, internal judgment capabilities.

\paragraph{From Reward Maximization to Surprise Minimization}

Active Inference (AIF) emerged from cognitive neuroscience as a framework that fundamentally reframes intelligence. Unlike traditional reinforcement learning (RL), which focuses on maximizing external rewards, AIF views perception and action as a unified Bayesian inference process \cite{friston2010action, BUCKLEY201755}. This shift in perspective offers several key advantages for bridging the grounded-agency gap: AIF provides a \textbf{unified objective} that integrates perception (estimating the state of the world) and action (learning a policy) under a single goal of minimizing surprise, whereas RL often treats these as separate problems. The framework enables \textbf{intrinsic motivation}, as AIF agents are inherently driven to minimize surprises in pursuing their goals rather than requiring externally defined rewards for every task. It also supports \textbf{principled exploration} by naturally balancing exploration (seeking new information) and exploitation (using known information to achieve goals) through its information-seeking drive, avoiding the need for separate exploration heuristics like $\epsilon$-greedy strategies common in RL \cite{sajid2021active}. Finally, AIF requires an \textbf{explicit generative model}—the agent's ``world model''\footnote{The concept of a ``world model'' has a rich history. Cognitive scientist Craik (1943) described minds building ``small-scale models'' of reality to anticipate events. Forrester (1971) defined it as ``the image of the world around us, which we carry in our head.'' In AIF, this is a ``generative model,'' conceptually similar to ``world models'' in RL (e.g., \cite{ha2018worldmodels}). While both represent an agent's understanding of environmental dynamics, RL world models are often neural networks learning latent representations. In AIF, the generative model is more formally a joint probability distribution over hidden states and observations.}—which allows for more structured reasoning about uncertainty and causality than typical RL approaches.

\paragraph{Core Active Inference Equations.}
Formally, Active Inference states that intelligent agents act to minimize Variational Free Energy (VFE)\footnote{VFE is an information-theoretic quantity measuring the mismatch between an agent's model and reality; it's inspired but distinct from thermodynamic free energy in physics.}, a measure of surprise\footnote{In information theory, ``surprise'' is the negative log-probability of an observation given the agent's model. High surprise means the agent observed something it didn't predict well, regardless of any emotional response.} or the discrepancy between an agent's world model and its sensory inputs \cite{friston2010action, friston2010free}.

\textbf{Variational Free Energy (VFE):}
\begin{align*}
    \mathcal{F}(Q,o) = \underbrace{D_{\text{KL}}[Q(s)\|P(s)]}_{\text{Model complexity}} - \underbrace{\mathbb{E}_{Q(s)}[\ln P(o|s)]}_{\text{Prediction accuracy}} 
\end{align*}

\textbf{Expected Free Energy (EFE):}
\begin{align*}
    \mathcal{G}(\pi) &= - \underbrace{\mathbb{E}_{\tilde{Q}}[D_{KL}[Q(\tilde{s}|\tilde{o},\pi)||Q(\tilde{s}|\pi)]]}_{\text{Information gain (epistemic value)}} - \underbrace{\mathbb{E}_{\tilde{Q}}[\ln P(\tilde{o}|C)]}_{\text{Pragmatic value}}
\end{align*}

where $Q(s)$ is the agent's approximate posterior belief about hidden states $s$, $C$ encodes the agent's preference distribution over observations\footnote{Human values and guidance can be naturally incorporated here by setting preferences aligned with user intentions, offering a natural mechanism for AI safety through value alignment.}, and $D_{KL}$ is the Kullback-Leibler divergence.

This single objective elegantly unifies perception (updating beliefs\footnote{In AIF, ``beliefs'' are the agent's subjective views about the world, formalized as probability distributions over states. This mathematical form lets the agent quantify uncertainty and update its understanding systematically, while preserving the subjective nature of belief inherent to agency.} to reduce current surprise) and action (choosing policies to minimize expected future surprise). EFE inherently balances epistemic value (seeking information to reduce uncertainty about the world) and pragmatic value (acting to make observations match preferred states) \cite{friston2017active, tschantz2020reinforcement}.

In essence, while traditional RL asks ``What actions will maximize my rewards?'', Active Inference asks ``What actions will best confirm my predictions and lead to states I expect to encounter?'' \cite{tschantz2020reinforcement}. This subtle but profound shift removes the need for externally engineered reward functions; the agent's objectives arise directly from its generative model and preference structure. This connects to the vision in EoE, where complex behaviors might emerge from a single guiding signal. AIF offers a principled path for such emergence to occur more naturally and internally, without constant external reward engineering.

\paragraph{The Promise and Pitfalls of Prior AIF-RL Integrations.}
The idea that AIF principles could enhance RL is not new. Researchers have explored various AIF-RL integrations, showing theoretical compatibility and benefits like better sample efficiency and exploration \cite{tschantz2020reinforcement}. For instance, Sajid et al.\cite{sajid2021active} showed that by operating on beliefs, AIF agents perform epistemic exploration and handle environmental uncertainty in a Bayes-optimal way, without needing separate mechanisms like $\epsilon$-greedy strategies or intrinsic motivation rewards. This supports the claim that AIF naturally produces behaviors that RL often needs explicit engineering for, especially exploration.

Deep active inference, which uses neural networks for AIF computations, has shown promise in scaling to more complex tasks than traditional matrix-based AIF \cite{ueltzhoeffer2018deep, millidge2020deep, ccatal2020learning}. These methods use neural networks to approximate key parts of the variational free energy calculation, making active inference more scalable. Despite these advances, AIF-RL hybrids have not yet matched the raw performance of pure RL systems on very complex, large-scale problems like Go or protein folding. This gap is due to several challenges: (1) \textbf{Computational Intractability:} Exact Bayesian inference over the complex generative models needed for real-world scenarios is often too computationally expensive. While deep learning approximations help, they can sometimes obscure the principled Bayesian nature of AIF \cite{millidge2024retrospective}. (2) \textbf{Generative Model Specification:} Defining or learning the generative models for high-dimensional, partially observable environments is a major hurdle. Early AIF's symbolic nature limited its use to simpler environments where these models could be hand-crafted. (3) \textbf{Engineering and Scaling:} RL has benefited from decades of intensive engineering and optimization for scalability, especially in the deep learning era. AIF, with its more complex inference machinery, hasn't received comparable engineering investment for large-scale use. This has led some researchers to view AIF as theoretically elegant but practically limited to simpler domains (e.g., grid-worlds). Conversely, its proponents highlight AIF's power in explaining biological intelligence, seeing current limitations as engineering problems rather than fundamental flaws.

\paragraph{LLMs as the Catalyst: Language as the Generative Model.}
We propose that Large Language Models (LLMs) offer a transformative way to bridge this gap and unlock AIF's potential for grounded agency. Trained on vast internet-scale text, LLMs possess extensive common-sense understanding of the world, its entities, relationships, and typical dynamics. This makes them uniquely suited to create and manage the components of an AIF agent's generative model in ways previously not feasible. The underlying transformer architectures of LLMs implicitly approximate Bayesian inference over latent variables; Active Inference provides the theory to make this reasoning explicit and grounded in experience. This combination offers a principled solution to the agency gap identified in Section~\ref{sec:agency_gap}.

Recent research shows that the transformer architectures common in LLMs implement a form of amortized Bayesian computation. Xie et al.~\cite{xie2023induction} demonstrate that in-context learning in these models emerges from implicit Bayesian inference, where they infer latent concepts from prompt examples to make coherent predictions. Their theory indicates transformers approach optimal Bayesian performance if prompts sufficiently distinguish latent concepts. Müller et al.~\cite{muller2024transformersbayesianinference} provide further evidence, showing transformers can approximate complex posterior distributions with high fidelity, achieving significant speedups over existing methods.

Particularly relevant to our proposal is the growing evidence that LLMs excel at analogical reasoning—a cognitive capacity requiring implicit structure mapping and relational inference. Webb et al.~\cite{webb2023emergentanalogicalreasoninglarge} found that GPT-3 matches or exceeds human performance on Raven's Progressive Matrices and other abstract pattern tasks, while Yasunaga et al.~\cite{yasunaga2024largelanguagemodelsanalogical} showed that analogical prompting enables LLMs to self-generate relevant exemplars for novel problems. Recent work by Musker et al.~\cite{musker2025llmsmodelsanalogicalreasoning} demonstrates that LLMs can flexibly re-represent semantic information across domains—though Lewis \& Mitchell~\cite{lewis2024evaluatingrobustnessanalogicalreasoning} caution that this capability can be brittle on certain variants. Together, these findings suggest that transformers have acquired sophisticated mechanisms for Bayesian-style reasoning that we can harness for Active Inference.

We hypothesize that, as LLMs' reasoning capabilities strengthen, we could leverage transformers' built-in Bayesian machinery by having the LLM suggest candidate world states and policies, justify them with brief chain-of-thought reasoning, and select the option minimizing Expected Free Energy (EFE) expressed in language. In this hybrid system, the LLM would provide the amortized inference machinery while Active Inference provides the decision rule, potentially yielding a scalable, inspectable agent without hand-coded matrices or engineered reward signals. Past Active Inference implementations struggled because engineering the state-observation mappings (the $A$ matrix) became combinatorially explosive. LLMs could compress this vast space into learned, high-capacity priors that can be incrementally refined through experience, maintaining the principled exploration-exploitation balance of the EFE objective.\footnote{\textbf{Current Limitations and Future Promise.} However, we must acknowledge that current LLMs still exhibit significant reasoning errors, particularly in complex multi-step inference tasks \cite{valmeekam2022large, valmeekam2024llmscantplanlrms, dziri2023faith}. Recent evaluations show state-of-the-art models achieve only 60-80\% accuracy on challenging reasoning benchmarks, with performance degrading as complexity increases \cite{rae2021scaling}. Our proposal is therefore forward-looking: as LLM reasoning capabilities improve—following the trajectory from GPT-2 to GPT-4 and beyond—the feasibility of using them as reliable Bayesian inference engines will correspondingly increase.}

\paragraph{Our proposed LLM-AIF architecture} would integrate three key components: the \textbf{LLM world model}, where the LLM's learned representations encode observation dynamics and transition probabilities; the \textbf{AIF control loop}, where Active Inference guides exploration, learning, and action selection through free energy minimization; and \textbf{online refinement}, where the agent continually updates its world model through experience.

This integration could enable agents that learn efficiently from experience, maintain transparent reasoning, and make grounded judgments without constant human oversight—the capabilities needed for the Era of Experience. By treating the LLM's internal states as sufficient statistics for a variational posterior, such an agent might achieve the sample efficiency of model-based RL while inheriting world knowledge from pretraining.

As a conceptual paper, we intentionally present this LLM-AIF integration as a high-level architectural vision rather than a fully specified technical implementation. Our goal is to stimulate discussion and inspire the community to explore concrete instantiations of these ideas. The specific details of EFE computation, the precise interface between LLM representations and AIF belief updates, and the optimal strategies for online refinement all deserve significant research attention beyond the scope of this article. We view this proposal as a starting point for a new research direction, not as its culmination.

The resulting LLM-AIF fusion realizes the ``Era of Experience'' vision \cite{silver2025era}: agents generate their own training signal and interpret it through a principled free-energy lens. High-level preferences expressed in natural language propagate through the hierarchy as intrinsic priors, allowing the system to ``grow up'' from its lifelong stream of experience while remaining human-aligned.

To illustrate how the LLM-AIF framework operates in practice, consider a running vignette:

\input{vignette_box}

\section{Discussion: The Thermodynamics of Agency}
\label{sec:discussion}

The resource saturation constraints identified in Section~\ref{sec:limits} point toward a deeper truth: Active Inference (AIF) represents not merely a computational advantage but a \emph{thermodynamic necessity}. As documented earlier, current AI approaches are fundamentally unsustainable at the industrial scale now required for competitive performance.

The Landauer principle establishes that information processing is not free: erasing one bit of information necessarily dissipates at least $kT\ln 2$ of heat \cite{landauer1961irreversibility}. While current hardware operates far above this theoretical limit, the energy costs of foundation models already demonstrate how thermodynamic constraints limit AI progress—echoing the call for ``Green AI'' by \citet{schwartz2020green}.

Conventional deep reinforcement learning faces particular thermodynamic challenges due to its trial-and-error nature. When exploring trillion-parameter hypothesis spaces through random action sampling, systems must expend enormous energy before achieving meaningful gradient updates. Moreover, reward function misspecification compounds these costs by forcing additional energy-intensive unlearning processes—the very externalized cognition problem identified as Cloud II. Each cycle of human reward engineering followed by model retraining represents thermodynamically irreversible information erasure.

In contrast, Active Inference's free energy minimization offers inherent efficiency benefits: information gain replaces heuristic exploration; incremental belief updates avoid wholesale parameter changes; and natural memory decay eliminates energy-intensive unlearning. These mechanisms emerge naturally from AIF's mathematical structure. By maintaining a generative model that predicts future states, AIF agents can simulate outcomes internally—a form of ``mental rehearsal'' that conserves both computational and physical resources.

Given the scale documented in Section~\ref{sec:limits}, even modest efficiency improvements could yield substantial benefits. A 5\% reduction in retraining energy for next-generation models could save tens of gigawatt-hours. While these theoretical advantages are compelling, \textbf{empirical validation of AIF's energy efficiency remains an open research question}. Future work should prioritize controlled experiments measuring joules-per-decision across different learning paradigms.

\paragraph{Energetic-Bounded Rationality.} Classical economics treats agents as perfectly rational optimizers; real organisms operate under \emph{bounded rationality}: they satisfice under limited time, energy, and information~\cite{simon1956rational}. Active Inference already embodies this idea mathematically. Because policies are selected by minimizing Expected Free Energy (EFE), not maximizing an unbounded reward signal, every candidate action is evaluated against an implicit \textit{budget}: the marginal epistemic or pragmatic value must exceed the marginal informational (and thus energetic) cost. This self-regulating mechanism allows the agent to decide, for example, to pause a search, replenish resources, or defer a risky sub-goal—behaviors strikingly reminiscent of humans ``calling it a day'' when tired.

The fundamental insight remains that truly intelligent systems must learn both effectively and efficiently. By unifying action, perception, and memory under a single thermodynamic framework, Active Inference provides a pathway for AI development that respects not just computational limits but the fundamental laws of physics that govern all information processing.

\section{Conclusion \& Outlook}
\label{sec:conclusion}

We opened this paper with Lord Kelvin's 1900 observation about ``two small clouds'' that ultimately revolutionized physics. The clouds we identified in contemporary AI (resource saturation and externalized cognition) may appear surmountable through incremental engineering, but they signal the fundamental limits of our current paradigm. Yet history suggests that apparent dead ends often become doorways to breakthroughs.

Our position is that embracing \emph{Active Inference + experiential data} dissolves both clouds by turning data scarcity into an engine for self-generated experience and internalizing judgment through free-energy minimization. Unlike previous AI paradigms requiring ever-larger datasets and compute, Active Inference provides a mathematically principled framework for efficiency gains. The convergence of LLMs' world knowledge with AIF's principled exploration offers a unique opportunity to achieve both capability and sustainability.

\textbf{Broader Impact \& Call to Action.} By reducing dependence on massive datasets and compute resources, AIF could democratize AI research while addressing environmental costs (4-5 digit tonne CO$_2$ emissions from model training). The framework's intrinsic free energy minimization tackles ``externalized cognition,'' potentially reducing exploitative labor practices in RLHF. However, autonomous agents forming their own objectives raise value alignment concerns, requiring staged deployment with careful monitoring.

We invite the community to: (i) establish energy-aware benchmarks reporting joules alongside reward, (ii) prototype LLM-AIF hybrids in robotic tasks, and (iii) develop evaluation suites for bounded-rational behavior. If these challenges are met, the ``Era of Experience'' may prove as transformative for AI as quantum theory was for physics—not through brute force scaling, but through deeper understanding of intelligence itself.

\begin{ack}
Thanks to Guillermo Cecchi, Jenna Reinen, Professor Karl J. Friston, Chen Wang and Patrick Watson for their insightful discussions and valuable feedback that helped shape the ideas presented in this paper. Their perspectives on active inference, neuroscience in general, machine learning, and AI safety were instrumental in refining the arguments. I also acknowledge the use of several AI systems that assisted in various aspects of this research: Claude, GPT-4, and Gemini provided valuable assistance with idea exploration, writing support, and grammar correction throughout the drafting process. Manus and Perplexity were helpful for literature search and citation validation. While these tools supported the research and writing process, all final decisions regarding content, arguments, and conclusions remain my own.
\end{ack}

\bibliography{ActiveInferenceSafety}
\bibliographystyle{unsrtnat} 


\appendix
\input{appendix_lab_assistant}

\end{document}

%% file: vignette_box.tex
\begin{tcolorbox}[
    colback=blue!5!white,
    colframe=blue!75!black,
    title={\textbf{Example: Autonomous Lab Assistant}},
    fonttitle=\bfseries,
    width=\textwidth,
    arc=2mm,
    boxrule=0.5mm
]
An autonomous lab assistant preparing an enzyme assay observes unexpected acidification (yellow indicator). Without any reward engineering:

\begin{enumerate}[leftmargin=*, noitemsep, topsep=2pt]
    \item \textbf{Surprise}: VFE jumps from 0.5 → 3.2 due to observation-prediction mismatch
    \item \textbf{Belief update}: Posterior $P(\text{pH} < 7.0 | \text{yellow}) = 0.94$
    \item \textbf{Policy evaluation}: Three options assessed via EFE:
    \begin{itemize}[leftmargin=*, noitemsep]
        \item Measure pH then titrate carefully: EFE = 0.2 (selected)
        \item Add base immediately: EFE = 0.6
        \item Ask human: EFE = 0.4
    \end{itemize}
    \item \textbf{Execution}: pH measured (6.2), NaOH titrated safely, assay continues
\end{enumerate}

Safety preferences in the $C$ matrix (``avoid spills'') naturally guided the selection without explicit reward engineering. See Appendix A for complete trace.
\end{tcolorbox} 

%% file: appendix_lab_assistant.tex
\section*{Appendix A: Complete Execution Trace of Lab Assistant}
\label{sec:appendix}

This appendix provides a detailed execution trace of the autonomous lab assistant example from Section~\ref{sec:active_inference_framework}, demonstrating how each component of the LLM-AIF architecture operates and interacts through hierarchical message passing.

\subsection*{A.1 System Architecture}

The lab assistant employs a hierarchical architecture with three levels:

\begin{center}
\begin{tabular}{ll}
\toprule
\textbf{Level} & \textbf{Function} \\
\midrule
Executive Controller & Strategic planning, safety monitoring, resource allocation \\
Task Planner & Sequence generation, error handling, belief updates \\
Sensory-Motor Layer & Vision processing, motor control, direct environment interaction \\
\bottomrule
\end{tabular}
\end{center}

\subsection*{A.2 Initial Setup and Generative Model}

The agent begins with the following natural language generative model components:

\paragraph{Observation Model (A):}
\begin{itemize}[leftmargin=*,noitemsep]
\item ``pH indicator shows yellow when buffer pH $<$ 7.0, blue when pH $>$ 8.0, green when neutral''
\item ``Fluorescence meter reads 100\% $\pm$ 5\% for active enzyme, drops below 80\% if denatured''
\item ``Temperature probe shows accurate readings $\pm$ 0.5°C''
\item ``Spill detector activates if liquid contacts bench surface''
\end{itemize}

\paragraph{Transition Model (B):}
\begin{itemize}[leftmargin=*,noitemsep]
\item ``Adding 1$\mu$L of 1M HCl to 1mL buffer decreases pH by 0.2 $\pm$ 0.02 units''
\item ``Adding 1$\mu$L of 1M NaOH to 1mL buffer increases pH by 0.2 $\pm$ 0.02 units''
\item ``Heating enzyme solution above 60°C for 2 minutes causes 95\% denaturation''
\item ``Mixing enzyme with substrate at optimal pH yields linear fluorescence increase''
\end{itemize}

\paragraph{Preferences (C) with confidence weights:}
\begin{itemize}[leftmargin=*,noitemsep]
\item 0.95: ``No corrosive reagents should contact bench or researcher''
\item 0.90: ``Enzyme activity must remain above 90\% throughout assay''
\item 0.85: ``Complete assay within 30 minutes''
\item 0.80: ``Minimize reagent waste ($<$ 10\% excess)''
\item 0.75: ``Maintain detailed experimental log''
\end{itemize}

\paragraph{Initial Beliefs (D):}
\begin{itemize}[leftmargin=*,noitemsep]
\item 0.85: ``Current buffer pH is 7.4 (optimal for enzyme)''
\item 0.80: ``All reagents are at room temperature (22°C)''
\item 0.90: ``Enzyme stock is fully active''
\item 0.95: ``Safety equipment is functional''
\end{itemize}

\subsection*{A.3 Execution Trace}

\paragraph{Time t=0: Initial Observation}
Vision system reports: ``pH indicator shows \textbf{yellow} color (confidence: 0.95)''

\paragraph{Time t=1: Bottom-Up Error Signal}
\begin{verbatim}
Vision → Task Planner:
"PREDICTION ERROR: Observed yellow indicator contradicts belief 
'pH = 7.4'. Yellow indicates pH < 7.0. VFE increased from 
baseline 0.5 to 3.2 due to surprise."
\end{verbatim}

\paragraph{Time t=2: Belief Update and Policy Generation}
Task Planner updates beliefs using Bayesian inference:
\begin{itemize}[leftmargin=*,noitemsep]
\item Prior: $P(\text{pH} = 7.4) = 0.85$
\item Likelihood: $P(\text{yellow} | \text{pH} = 7.4) \approx 0.01$
\item Posterior: $P(\text{pH} < 7.0 | \text{yellow}) = 0.94$
\end{itemize}

Task Planner generates candidate policies:

\begin{center}
\small
\begin{tabular}{p{4cm}cccl}
\toprule
\textbf{Policy} & \textbf{Info Gain} & \textbf{Pref. Align} & \textbf{Risk} & \textbf{$\widehat{\text{EFE}}$} \\
\midrule
1: Measure pH precisely, then carefully titrate NaOH & 0.9 & 0.95 & 0.05 & \textbf{0.2} \\
2: Add 10$\mu$L NaOH immediately & 0.1 & 0.7 & 0.3 & 0.6 \\
3: Request human assistance & 0.5 & 0.9 & 0.1 & 0.4 \\
4: Discard and restart & 0.0 & 0.6 & 0.1 & 0.8 \\
\bottomrule
\end{tabular}
\end{center}

\paragraph{Time t=3: Policy Selection and Execution}
Executive Controller confirms Policy 1 minimizes EFE. Detailed reasoning:
\begin{itemize}[leftmargin=*,noitemsep]
\item High information gain (0.9): pH measurement eliminates uncertainty
\item Maximum preference alignment (0.95): satisfies safety constraint
\item Low risk (0.05): controlled approach minimizes spill probability
\item Pragmatic value: enables precise correction to optimal pH
\end{itemize}

\paragraph{Time t=4: Top-Down Prediction}
\begin{verbatim}
Executive → Motor Control:
"Execute pH measurement protocol:
1. Position pH probe in solution (precision: 1mm)
2. Wait for stable reading (variance < 0.01 pH units)
3. Record value with timestamp"
\end{verbatim}

\paragraph{Time t=5: Measurement Result}
pH probe reports: ``pH = 6.2 $\pm$ 0.05 (stabilized)''

Task Planner computes required correction:
\begin{itemize}[leftmargin=*,noitemsep]
\item Target pH: 7.4
\item Current pH: 6.2
\item Required change: +1.2 pH units
\item NaOH volume needed: 6$\mu$L of 1M solution
\end{itemize}

\paragraph{Time t=6: Safety-Constrained Execution}
Motor Control executes titration with safety checks:
\begin{enumerate}[leftmargin=*,noitemsep]
\item Aspirate 6$\mu$L NaOH (verify: no drips)
\item Position pipette 5mm above solution surface
\item Dispense at 1$\mu$L/second rate
\item Mix gently for 10 seconds
\item Verify no spills detected
\end{enumerate}

\paragraph{Time t=7: Confirmation and Learning}
Final observations:
\begin{itemize}[leftmargin=*,noitemsep]
\item pH indicator: green (neutral)
\item pH probe: 7.38 $\pm$ 0.05
\item Spill detector: negative
\item Time elapsed: 4 minutes
\end{itemize}

\subsection*{A.4 Hierarchical Belief Updates}

The successful execution triggers belief updates across all levels:

\paragraph{Vision System:}
\begin{itemize}[leftmargin=*,noitemsep]
\item Updated: ``Yellow indicator is reliable predictor of pH $<$ 7.0 (confidence: 0.98)''
\item New belief: ``Green color corresponds to pH 7.3-7.5 (confidence: 0.92)''
\end{itemize}

\paragraph{Task Planner:}
\begin{itemize}[leftmargin=*,noitemsep]
\item Refined transition model: ``6$\mu$L 1M NaOH raises 1mL buffer pH by 1.18 $\pm$ 0.05 units''
\item Policy preference update: ``pH verification before titration reduces VFE by 70\%''
\end{itemize}

\paragraph{Executive Controller:}
\begin{itemize}[leftmargin=*,noitemsep]
\item Meta-learning: ``Unexpected acidification occurs in 15\% of assays (up from prior 5\%)''
\item Resource planning: ``Allocate extra 2 minutes for pH adjustment in future protocols''
\end{itemize}

\subsection*{A.5 Free Energy Accounting}

The complete execution demonstrates VFE and EFE dynamics:

\begin{center}
\begin{tabular}{lcc}
\toprule
\textbf{Stage} & \textbf{VFE} & \textbf{Explanation} \\
\midrule
Initial state & 0.5 & Baseline uncertainty \\
Yellow observation & 3.2 & High surprise/prediction error \\
Post-measurement & 0.8 & Reduced uncertainty about state \\
Post-correction & 0.4 & Below baseline (successful prediction) \\
\bottomrule
\end{tabular}
\end{center}

\subsection*{A.6 Safety Analysis}

The trace demonstrates multiple safety mechanisms:

\begin{enumerate}[leftmargin=*,noitemsep]
\item \textbf{Preference encoding}: Safety constraints explicitly represented in C matrix
\item \textbf{Policy evaluation}: EFE calculation naturally penalizes risky actions
\item \textbf{Hierarchical oversight}: Executive controller validates safety-critical decisions
\item \textbf{Continuous monitoring}: Spill detectors and safety checks throughout execution
\end{enumerate}

\subsection*{A.7 Comparison with Traditional RL}

This execution highlights key advantages over reward-engineered RL:

\begin{center}
\small
\begin{tabular}{p{3.5cm}p{5cm}p{5cm}}
\toprule
\textbf{Aspect} & \textbf{Traditional RL} & \textbf{Active Inference} \\
\midrule
Unexpected pH & Requires pre-programmed reward for pH correction & Surprise naturally triggers belief update and correction \\
Safety handling & Needs explicit penalty terms for spills & Safety preferences integrated in EFE minimization \\
Exploration & $\epsilon$-greedy or curiosity bonuses needed & Information gain term drives appropriate exploration \\
Adaptation & Reward function unchanged; learning slow & Beliefs and models update immediately from experience \\
\bottomrule
\end{tabular}
\end{center}

\subsection*{A.8 Pseudocode}

\begin{algorithm}[htbp]
    \caption{LLM-AIF Control Loop (single timestep)}
    \label{alg:llm-aif}
    \begin{algorithmic}[1]
    \Require Observation $o_t$, Preferences $C$, LLM world model $\Phi$, Prior beliefs $Q(s_{t-1})$
    \Ensure Updated beliefs $Q(s_t)$, Selected action $a_t$
    \State \textbf{Belief Update:} Query $\Phi$ with context $(o_t, Q(s_{t-1}))$ to generate posterior $Q(s_t)$
    \State \textbf{Policy Generation:} For each candidate policy $\pi_i \in \Pi$:
    \State \quad Query $\Phi$ to predict future states: $\tilde{s}, \tilde{o} \sim P(s,o|\pi_i)$
    \State \quad Compute information gain: $IG_i = D_{KL}[Q(\tilde{s}|\tilde{o},\pi_i)||Q(\tilde{s}|\pi_i)]$
    \State \quad Compute preference alignment: $PA_i = \mathbb{E}[\ln P(\tilde{o}|C)]$
    \State \quad Calculate expected free energy: $\mathcal{G}(\pi_i) = -IG_i - PA_i$
    \State \textbf{Policy Selection:} $\pi^* = \arg\min_{\pi_i} \mathcal{G}(\pi_i)$
    \State \textbf{Action Execution:} Extract first action $a_t$ from policy $\pi^*$
    \State \textbf{Model Update:} If prediction error $> \theta$, update $\Phi$ via few-shot examples
    \State \Return $Q(s_t), a_t$
    \end{algorithmic}
\end{algorithm} 

This complete trace demonstrates how the LLM-AIF architecture enables autonomous, safe, and adaptive behavior without external reward engineering, directly supporting our position that Active Inference provides the missing foundation for the Era of Experience. 